\pdfoutput=1
\documentclass[11pt]{article}

\usepackage[preprint]{acl}
\usepackage{pifont}
\usepackage{amssymb}
\usepackage{subcaption}
\usepackage{booktabs}
\usepackage{times}
\usepackage{latexsym}
\usepackage{algorithm}
\usepackage{algpseudocode}
\usepackage[T1]{fontenc}
\usepackage[utf8]{inputenc}

\usepackage{microtype}
\usepackage{listings}
\usepackage{tcolorbox}
\usepackage{graphicx}

\usepackage{inconsolata}
\usepackage{amsmath}

\title{ORPP: Self-Optimizing Role-playing Prompts to Enhance Language Model Capabilities}

\author{Yifan Duan, Yihong Tang, \\ {\bf Kehai Chen\footnotemark[2], Liqiang Nie, Min Zhang} \\
        Harbin Institute of Technology, Shenzhen, China \\
        \texttt{\{ziye\_0926@outlook.com,  chenkehai@hit.edu.cn\}}
}

\begin{document}
\maketitle
\begin{abstract}

\begingroup\def\thefootnote{\dag}\footnotetext{Corresponding author.}\endgroup
High-quality prompts are crucial for eliciting outstanding performance from large language models (LLMs) on complex tasks. Existing research has explored model-driven strategies for prompt optimization. However, these methods often suffer from high computational overhead or require strong optimization capabilities from the model itself, which limits their broad applicability.
To address these challenges, we propose ORPP (Optimized Role-Playing Prompt),a framework that enhances model performance by optimizing and generating role-playing prompts. The core idea of ORPP is to confine the prompt search space to role-playing scenarios, thereby fully activating the model's intrinsic capabilities through carefully crafted, high-quality role-playing prompts. Specifically, ORPP first performs iterative optimization on a small subset of training samples to generate high-quality role-playing prompts. Then, leveraging the model’s few-shot learning capability, it transfers the optimization experience to efficiently generate suitable prompts for the remaining samples.
Our experimental results show that ORPP not only matches but in most cases surpasses existing mainstream prompt optimization methods in terms of performance. Notably, ORPP demonstrates superior “plug-and-play” capability. In most cases, it can be integrated with various other prompt methods and further enhance their effectiveness.
\end{abstract}

\section{Introduction}
Large language models (LLMs) have demonstrated outstanding performance in handling various complex tasks~\cite{Marco2024PronVP,li-lan-2025-large,Son2024MultiTaskIC}. To fully unleash the capabilities of these models, carefully designed prompts have been extremely important~\cite{Wu2024StraGoHS}. Research has shown that specific prompt strategies can significantly influence the depth of reasoning and creativity of models in particular tasks~\cite{Xiang2025SelfSupervisedPO}. However, manually designing prompts is challenging and labor-intensive. As a result, researchers have begun to explore leveraging the capabilities of LLMs themselves to automatically generate prompts, opening up new avenues for the field of prompt engineering.
\begin{figure}[t]
    \centering
    \includegraphics[width=0.5\textwidth]{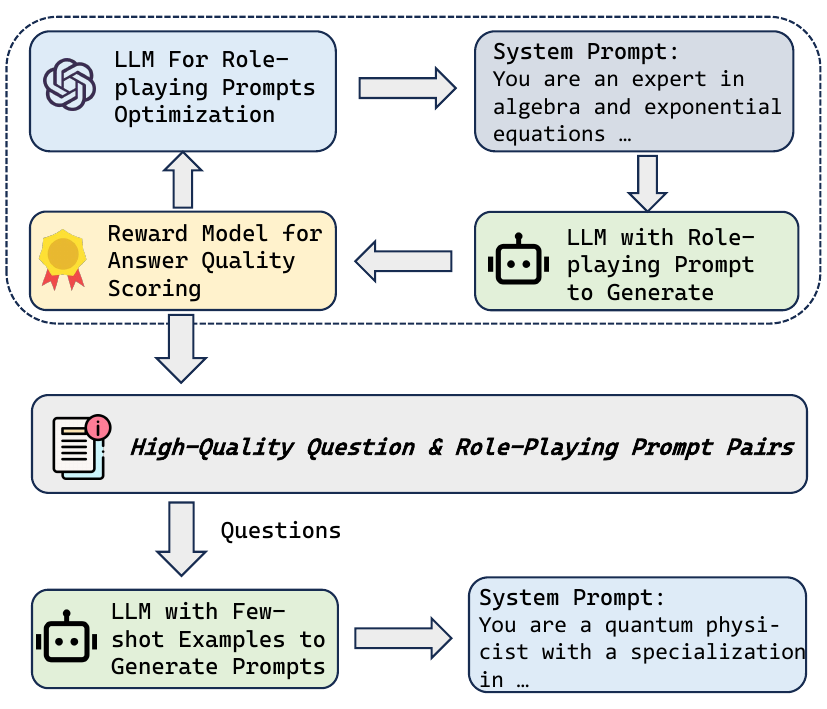}
    \caption{ORPP: A two-stage framework for generating and optimizing role-playing prompts, combining iterative optimization and few-shot learning.}
    \label{intro}
\end{figure}

Although techniques such as iterative optimization, textual gradient, and self-supervised prompt optimization have shown the potential of LLMs to design high-quality prompts, these methods still face significant challenges~\cite{Yuksekgonul2024TextGradA,wang2024promptagent,Xiang2025SelfSupervisedPO}. For example, iterative optimization methods often require large amounts of data to evaluate prompt quality, resulting in high computational costs. Self-supervised prompt optimization relies on the intrinsic evaluation capabilities of LLMs, which significantly reduces costs but places higher demands on the performance of the LLM itself~\cite{Xiang2025SelfSupervisedPO}. In addition, searching within the vast prompt space further increases the difficulty of discovering high-quality prompts.

Previous studies have shown that by setting specific role-playing prompts, the reasoning ability and creativity of LLMs can be enhanced, thereby improving their performance in both closed and open-ended tasks~\cite{Chen2024HoLLMwoodUT,kong-etal-2024-better,Lu2024LLMDE}. For example, guiding the model to play the role of an expert in a specific professional field, or simulating a character with a particular way of thinking, can prompt the model to examine problems from multiple perspectives. This finding reveals a way for LLMs to further stimulate their potential by simulating human roles. However, existing methods mostly rely on manually preset role-playing prompts, making it difficult to automatically generate high-quality corresponding role settings according to specific task requirements, which limits the possibility of further improving model performance through automatic role simulation.

Based on the above research insights, this paper proposes a novel role-playing-based prompt optimization method. Unlike existing methods, our approach focuses on role-playing prompts and aims to generate high-quality role-playing prompts to enhance the reasoning ability of large language models. This method can automatically generate optimized role-playing prompts for specific tasks, effectively stimulating the model’s potential in particular task scenarios through role-playing. In addition, as a plug-and-play modular component, our method can be combined with existing prompt engineering techniques to further improve the performance of other methods.

Experimental results show that our method significantly enhances model performance in closed tasks by generating high-quality role-playing prompts. Furthermore, experiments verify the effectiveness of our method as a plugin, demonstrating that it can work synergistically with various existing methods and further improve their overall performance in most cases. Our exploratory experiments reveal that role-playing prompts generated by smaller-parameter models can be successfully transferred to larger-parameter models and exhibit excellent performance, further demonstrating the generalization potential of our prompt generation method.

Our main contributions are as follows:  
\begin{enumerate}
    \item We propose a novel role-playing-based prompt optimization framework, which enhances the performance of large language models on specific tasks by automatically generating high-quality role-playing prompts.
    \item We show that our method works as a plug-and-play module and can be combined with different prompt engineering techniques. In most cases, it helps these techniques achieve even better results.
    \item By focusing on role-playing prompts, our approach reduces the difficulty of prompt optimization, and the generated prompts demonstrate strong transferability.
\end{enumerate}

\section{Related Work}
\paragraph{Role-Playing} Large language models (LLMs) have demonstrated great potential in the field of role-playing~\cite{Lu2024LargeLM,Chuang2024BeyondDA,TangWZJZHH24}. Through carefully designed prompts or targeted fine-tuning, LLMs can simulate the language style, personality traits, and knowledge systems of specific characters, successfully portraying a wide range of roles from historical figures to fictional literary characters~\cite{Chen2024FromPT,shao2023characterllm}.
Further research has revealed that endowing LLMs with specific roles can unlock deeper capabilities of the models. Appropriately designed role-playing prompts have been proven to effectively enhance the model’s creativity in open-ended tasks and improve its performance in closed tasks ~\cite{kong2024betterzeroshotreasoningroleplay,Lu2024LLMDE,DBLP:journals/corr/abs-2407-08995}.
However, current approaches often rely on manually crafted role-playing prompts or use fine-tuned models to generate prompts. The models themselves struggle to autonomously generate high-quality role-playing prompts that can significantly improve task performance.
Our work employs automatic prompt optimization techniques, enabling the model to generate high-quality role-playing prompts for itself that are tailored to specific tasks.

\paragraph{Prompt Optimization} Carefully crafted prompts are crucial for optimizing the performance of large language models (LLMs)~\cite{Wang2024PromptEI,Wei2022ChainOT,wang2024promptagent}. However, creating effective prompts often requires in-depth task-specific knowledge. In recent years, researchers have begun to explore leveraging the capabilities of LLMs themselves to automatically generate high-quality prompts~\cite{Fernando2023PromptbreederSS,wang2024promptagent,Gao2025ThePA}. This strategy not only reduces reliance on human expertise but also enables the dynamic generation of more targeted prompts according to specific task requirements, with the PO method demonstrating great potential in this process.
However, when dealing with complex tasks, relying on LLMs to generate high-quality prompts usually requires the model to possess strong comprehension abilities~\cite{Wang2024ExploringPP}. At the same time, searching for high-quality prompts within an unrestricted prompt space further increases the difficulty and cost of optimization.Our work proposes optimizing role-playing prompts to further unlock the potential of LLMs~\cite{Pryzant2023AutomaticPO}. By constraining the optimization space of prompts, role-playing prompts reduce the difficulty of generating high-quality prompts and can be flexibly integrated as plugin modules into existing research frameworks.
\section{Method}
Our method is divided into two stages: First, through iterative optimization, we find high-quality role-playing prompts for sample data that can effectively elicit the model's capabilities. Then, these optimal prompts are used as few-shot examples to guide the model in generating corresponding role prompts for other problems.
\begin{figure*}[t]
    \centering
    \includegraphics[width=0.98\textwidth]{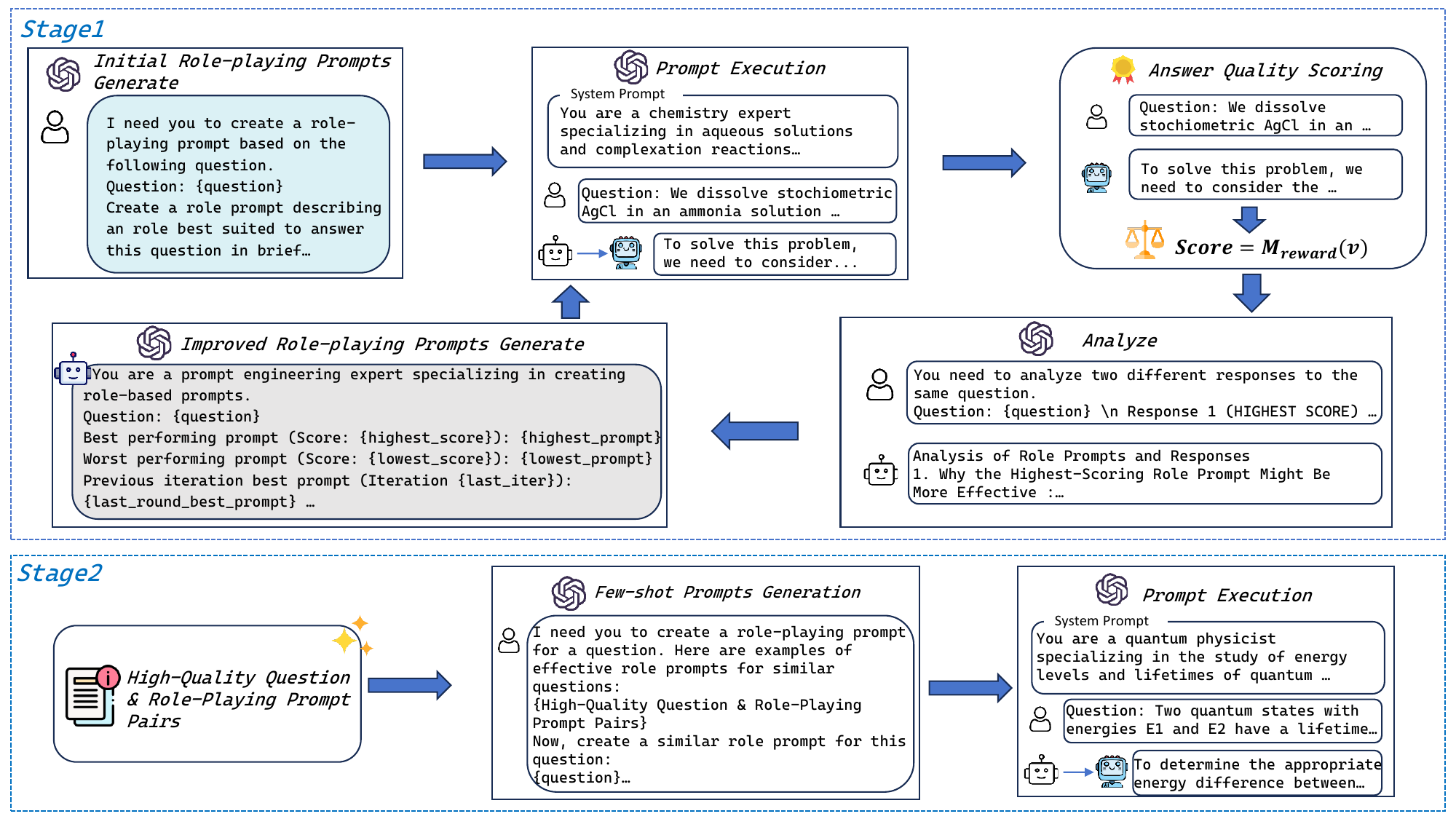}
    \caption{Illustration of ORPP: A two-stage role-playing prompt generation and optimization framework that combines iterative optimization and few-shot learning. Using a real example from GPQA, in Stage One, high-quality role-playing prompts are identified for sample data through an iterative optimization process. In Stage Two, these optimized High-Quality Question \& Role-Playing Prompt Pairs are used as few-shot examples to guide the model in generating corresponding role-playing prompts for new questions, thereby efficiently improving the model’s performance on a wide range of questions.}
    \label{main}
\end{figure*}

\subsection{Optimization Objective}
Role-playing prompt optimization aims to automatically generate the optimal role-playing prompt R for an input question Q. Its core objective is to set the most effective role prompt for the model, thereby maximizing the model's performance on that question. This optimization objective can be represented as: 
$$R^* = \underset{R \in \mathcal{R}}{\arg\max}\ \mathcal{S}(\mathcal{M}(Q | R))$$ 
where \(\mathcal{R}\) represents the space of candidate role-playing prompts, the scale of which is significantly reduced compared to traditional prompt optimization. \(\mathcal{M}(Q | R)\) denotes the output generated by model M for question Q given the role-playing prompt R. And \(\mathcal{S}\) is the evaluation function used to quantify the quality of the model's output.

\subsection{Optimization Process}
The optimization process first generates role-playing prompts for the extracted samples, and then iteratively refines them to ensure these prompts meet high-quality standards. The resulting pairs of high-quality samples and prompts will be used for subsequent few-shot transfer learning.

As shown in Algorithm \ref{alg:role_prompt_simple_revised}, the optimization process mainly relies on the model's own reasoning and generation capabilities to achieve automatic iteration and improvement of prompts. Previous studies have shown that the quality of prompts is reflected in the performance of the model's output~\cite{Deng2023RephraseAR,Xiang2025SelfSupervisedPO}. Based on this, our method relies on evaluating the quality of the model's output to optimize the prompts.This is accomplished through using a reward model.

First, a small sample set \(Q\) is selected from the dataset. For each question \(q \in Q\), the model \(\mathcal{M}\) generates a baseline answer \(A_{\text{base}}(q) = \mathcal{M}(q)\) without any specific role prompt. Then, the reward model \(\mathcal{M}_{\text{reward}}\) evaluates this baseline answer to obtain a baseline score \(\mathcal{S}(q) = \mathcal{M}_{\text{reward}}(q, A_{\text{base}}(q))\), which serves as a reference standard for measuring the effectiveness of subsequent role-playing prompts.

During the optimization phase, for each question \(q\), the model first generates \(k\) initial candidate role-playing prompts to form the initial prompt set \(R_0(q)\). Then, \(N\) rounds of iterative optimization are performed. In each round \(i\), the model uses each prompt \(R \in R_{i-1}(q)\) from the current prompt set to generate an answer \(\mathcal{M}(q|R)\). The reward model \(\mathcal{M}_{\text{reward}}\) then calculates the score \(\mathcal{S}({M}(q|R))\). Throughout all rounds, the system continuously records the highest-scoring prompt found so far for question \(q\), denoted \(R_{\text{best}}(q)\), and the lowest-scoring prompt, \(R_{\text{worst}}(q)\), updating them as better or worse prompts are discovered in each round's evaluations.

In the analysis and improvement phase of each iteration \(i\), where the goal is to generate the new prompt set \(R_i(q)\) based on the evaluation of \(R_{i-1}(q)\): The model receives comprehensive feedback. This includes the globally best-performing prompt \(R_{\text{best}}(q)\) and globally worst-performing prompt \(R_{\text{worst}}(q)\) identified across all rounds up to and including the evaluation of the set \(R_{i-1}(q)\), along with their corresponding answers and scores. Crucially, to avoid falling into cycles or local optima, the model also considers the global best prompt from the end of the immediately preceding iteration, \(R_{\text{best}}^{\text{prev\_iter}}(q)\) .

Based on this rich feedback, the model analyzes the reasons for performance differences. This analytical process can be conceptualized as estimating the effective "text gradient" of the reward function \(\mathcal{S}\) with respect to the prompt \(R\), denoted as \(\mathcal{A}^{(i-1)}=\nabla_{R} \mathcal{S}(R_{\text{ref}}^{(i-1)}; q)\). This gradient provides directional guidance for prompt optimization. Here, \(R_{\text{ref}}^{(i-1)}\) represents the reference prompt(s) from the evaluation of \(R_{i-1}(q)\). \(\mathcal{A}^{(i-1)}\) contains the full analytical context of the evaluation.

Based on this "gradient", the system formulates targeted improvement suggestions. Following these suggestions, the model generates a new set of \(k\) candidate role-playing prompts, \(\{R_{j}^{(i)}(q) | j=1, \dots, k\}\), which collectively form the next prompt set \(R_i(q)\). The generation of each new candidate prompt \(R_{j}^{(i)}(q)\) can be described by the following transformation:

\[
R_{j}^{(i)}(q) = \mathcal{T}_{\text{gen}} \left( R_{\text{set}}^{(i-1)}(q), \nabla_{R} \mathcal{S}, T \right)
\]

where \(R_{\text{set}}^{(i-1)}(q)\) is the prompt set, based on the current global best prompt \(R_{\text{best}}(q)\), the current global worst prompt, and the previous round’s best prompt;\(\mathcal{T}_{\text{gen}}\) is a model-based generative function used to generate new role-playing prompts. \(T\) is the temperature set during model generation.

After \(N\) rounds of iteration, for each question \(q\), the prompt with the highest score recorded across all rounds, \(R_{\text{best}}(q)\), which is equivalent to \(R_{\text{best}}^{(N)}(q)\) at this stage, is selected as the final optimized result \(R^*(q)\) for that question.
\begin{algorithm}[t]
\caption{Simplified Optimization Process Of ORPP}
\label{alg:role_prompt_simple_revised}
\begin{algorithmic}[1]

\ForAll{$q \in Q$}
    \State $A_{base}(q) \gets \mathcal{M}(q)$; 
    \State $\mathcal{S}_{base}(q) \gets \mathcal{M}_{reward}(q, A_{base}(q))$

    \For{$i = 0$ to $N-1$}
            \State $R_{set}(q) \gets \mathcal{G}(\mathcal{M}, q)$ 

        \ForAll{$R \in R_{set}(q)$} 
            \State $A(q, R) \gets \mathcal{M}(q | R)$; 
            \State $S_{current} \gets \mathcal{M}_{reward}(q, A(q, R))$ 

            \If{$S_{current} > \mathcal{S}_{best}(q)$} 
                \State $\mathcal{S}_{best}(q) \gets S_{current}$; $R_{best}(q) \gets R$
            \EndIf
             \If{$S_{current} < \mathcal{S}_{worst}(q)$} 
                \State $\mathcal{S}_{worst}(q) \gets S_{current}$; $R_{worst}(q) \gets R$
            \EndIf
        \EndFor
    \EndFor
        \State $R^*(q) \gets R_{best}(q)$
\EndFor

\end{algorithmic}
\end{algorithm}
\subsection{Prompts Generation}

After identifying a high-quality role-playing prompt $R^*(q)$ for each question $q$ in the randomly selected training subset during the optimization phase, we adopt a few-shot approach to generalize this optimization capability to test questions, as shown in Algorithm \ref{alg:few_shot_generation_test}.

Specifically, we first select the top $m$ highest-quality "question and optimal role-playing prompt" pairs $(q_{train}, R^*(q_{train}))$ from the optimized samples to construct a few-shot example set $E_{fs}$. The quality of each pair is evaluated based on the improvement in answer score when using the role-playing prompt compared to answering without one.

Next, for each new question $q$ in the test set $Q_{test}$, the model $\mathcal{M}$ receives the new question $q$ along with the few-shot example set $E_{fs}$ as input. Through few-shot learning, the model can understand the association patterns between questions and prompts in the examples, and based on this understanding, infer and generate an appropriate role-playing prompt $R_{fs}(q)$ for the current test question. This process leverages the strong in-context learning ability of large language models, enabling them to mimic the patterns in the examples and produce high-quality role-playing prompts.

Finally, the generated role-playing prompt $R_{fs}(q)$ is used as the system prompt, and together with the test question $q$ to be inputed into the model to obtain the final answer $A_{final}$ for the test question. This method avoids time-consuming multi-round optimization for each question, significantly improves efficiency, and effectively transfers the optimization strategies learned from the training subset to new questions.

\begin{algorithm}[h]
\caption{Few-Shot Role Prompt Generation and Final Answer Generation for Test Set} 
\label{alg:few_shot_generation_test} 
\begin{algorithmic}[1] 

\State Select $m$ pairs $(q_{train}, R^*(q_{train}))$ where $q_{train} \in Q_{sub}$ to form the few-shot example set $E_{fs} = \{(q_1, R^*(q_1)), \dots, (q_m, R^*(q_m))\}$. 

\ForAll{$q \in Q_{test}$} 
    \State $R_{fs}(q) \gets \mathcal{G}_{fs}(\mathcal{M}, q, E_{fs})$ 
\State $A(q, R_{fs}(q))$ $\gets$ 
$\mathcal{M}(q | R_{fs}(q))$  
    \State $A_{final} \gets A(q, R_{fs}(q))$ 
\EndFor

\end{algorithmic}
\end{algorithm}

\section{Experiments}
  \subsection{Experimental Setup}
  \paragraph{Benchmarks} 
  We select a diverse set of tasks to evaluate the effectiveness of our method in enhancing model capabilities and to verify its performance when combined with other methods as a plugin. Specifically, we choose several tasks that rely on model reasoning abilities, including GPQA~\cite{Rein2023GPQAAG}, MMLU-CF~\cite{Zhao2024MMLUCFAC}, MMLU-Pro~\cite{Wang2024MMLUProAM}, MATH~\cite{hendrycksmath2021}, AGIEval-Math~\cite{Zhong2023AGIEvalAH}, and Med QA~\cite{jin2021disease}. For GPQA, we use the most challenging GPQA-diamond subset as the test set, with the remaining questions as the training set. In MMLU-CF and MMLU-Pro, we select tasks from four categories: mathematics, physics, chemistry, and biology. Specifically, we randomly sample 50 data points from each category as the training set, and use the remaining data as the test set. For MATH, we use the highest difficulty level, level 5, as the test set. In AGIEval-Math, we use questions with difficulty levels 4 and 5 as the test set, with the rest as the training set. For Med QA, we use the entire test set for evaluation. In Table \ref{tab:dataset_sizes}, we present the sizes of the training and test sets for each dataset we used. Details of the dataset construction are provided in Appendix~\ref{Dataset}.

\begin{table}[h!] % [h!] suggests placing the table here if possible
\centering % Center the table
\captionsetup{justification=centering} % Center the caption (requires caption package)

\begin{tabular}{lll} % l: left-align, r: right-align
\toprule % Top thick rule
Dataset & Training Set& Test Set\\
\midrule % Middle standard rule
GPQA & 250&198 \\
MMLU-CF &150 &2528 \\
MMLU-Pro &150 &4299 \\
MATH &7500 &1324 \\
AGIEval-Math &512 &488 \\
Med QA &10178 &1273 \\
\bottomrule % Bottom thick rule
\end{tabular}
\caption{Training and Test Set Sizes of Datasets}
\label{tab:dataset_sizes}
\end{table}

\paragraph{Comparison Methods} To evaluate the effectiveness of our proposed role-playing-based prompt optimization method in enhancing the reasoning capabilities of large language models, we compare it with various advanced prompt engineering techniques. In this study, direct question input was set as the baseline. In the controlled experiments, we select several classic prompting methods, including Chain-of-Thought ~\cite{Wei2022ChainOT}, Rephrase ~\cite{Deng2023RephraseAR}, and Step-Back ~\cite{zheng2024take}. In addition, our method was also compared with current state-of-the-art prompt optimization approaches such as OPRO~\cite{yang2024large} and SPO~\cite{Xiang2025SelfSupervisedPO}. Detailed experimental configurations for each comparison method can be found in the Appendix~\ref{Implement}. To further confirm the practical application potential of our method as a plugin, we also explore its combination with the aforementioned techniques and conduct relevant studies on these combinations.

\paragraph{Implementation Details} This study selects two models of different scales, Qwen2.5-14B and Qwen2.5-32B~\cite{qwen2.5}, for experimental validation. All models are deployed using vLLM~\cite{kwon2023efficient}, and inference is performed on NVIDIA L20 GPUs. To ensure the reproducibility of the experiments, we set a fixed random seed of 42 and use it to sample initial instances from the training set. During the prompt optimization phase, we randomly select ten samples from the training set for optimization. In this process, we employ ArmoRM-Llama3-8B-v0.1~\cite{Wang2024InterpretablePV} as the reward model to evaluate the quality of the generated content. The model is deployed using Hugging Face Transformers~\cite{Wolf2019HuggingFacesTS}. During the optimization phase, we will conduct ten rounds of iteration, generating three role-playing prompts in each round. In the second stage, we set the number of few-shot examples to 3 and further investigate the impact of different numbers of examples on the final model performance in subsequent experiments.

Regarding the temperature parameter, we set it to 0.2 only when generating role-playing prompts to introduce moderate diversity. In all other generation and evaluation steps, the temperature is set to 0 to ensure deterministic outputs. In the subsequent transfer application stage, in order to maintain consistency and reproducibility, the temperature remains 0 for both generating role-playing prompts and producing final answers with the model. Detailed experimental settings for the comparative methods are presented in the appendix.

\subsection{Experiments Results}
\paragraph{Main Results} In Table \ref{tab:model_comparison_grouped}, we present the performance of our method compared with other approaches on various benchmark tests. It shows that our method achieves outstanding results in most tasks, especially on multiple tasks such as GPQA, AGIEval-Math, and MMLU-PRO, where it achieves the best performance on its own. This fully demonstrates the effectiveness of high-quality role-playing prompts in improving model performance and validates the effectiveness of our approach. It is worth emphasizing that, before being used as a plugin, our method alone can effectively enhance the model’s performance and achieve results comparable to advanced prompt-based methods.

\begin{table*}[t]
 \centering

 \resizebox{\textwidth}{!} % 调整表格宽度以适应文本宽度
 {
 \begin{tabular}{lcccccc} % 移除之前的 Model 列，因为现在按模型分组
 \toprule
 Method & GPQA & AGIEval-MATH & MATH & MMLU-Pro & MMLU-CF & Med QA \\
 \midrule
 \multicolumn{7}{c}{Qwen2.5-14B} \\ % 添加 qwen14B 分组标题
 \midrule
 base & 43.43 & 67.21 &59.29 & 71.27 &70.02  & 63.63\\
 CoT  & 43.94 { (\textcolor{green!60!black}{+0.51})} & 68.24 { (\textcolor{green!60!black}{+1.03})} & 60.57 { (\textcolor{green!60!black}{+1.28})} &71.85 { (\textcolor{green!60!black}{+0.58})}  & 71.88 { (\textcolor{green!60!black}{+1.86})} & \textbf{66.69} { (\textcolor{green!60!black}{+3.06})}\\
 Rephrase  &40.40 { (\textcolor{red!80!black}{-3.03})}  &  68.24 { (\textcolor{green!60!black}{+1.03})}& 58.01 { (\textcolor{red!80!black}{-1.28})} & 72.27 { (\textcolor{green!60!black}{+1.00})} & \textbf{72.47} { (\textcolor{green!60!black}{+2.45})} & 66.06 { (\textcolor{green!60!black}{+2.43})} \\
 Step-back & 43.43 { (\textcolor{gray}{+0.00})} & 70.08 { (\textcolor{green!60!black}{+2.87})} &  57.85 { (\textcolor{red!80!black}{-1.44})}&72.18 { (\textcolor{green!60!black}{+0.91})}  & 72.19 { (\textcolor{green!60!black}{+2.17})}& 65.83 { (\textcolor{green!60!black}{+2.20})} \\
  OPRO & 43.94 { (\textcolor{green!60!black}{+0.51})} & 68.85 { (\textcolor{green!60!black}{+1.64})}& 61.78 { (\textcolor{green!60!black}{+2.49})}& 70.99 { (\textcolor{red!80!black}{-0.28})} & 71.24 { (\textcolor{green!60!black}{+1.22})} & 65.20 { (\textcolor{green!60!black}{+1.57})} \\
 SPO &42.93 { (\textcolor{red!80!black}{-0.50})}  &68.24 { (\textcolor{green!60!black}{+1.03})}  & 59.74 { (\textcolor{green!60!black}{+0.45})} &\textbf{73.34} { (\textcolor{green!60!black}{+2.07})}  &70.93 { (\textcolor{green!60!black}{+0.91})} &64.57 { (\textcolor{green!60!black}{+0.94})}\\
 ORPP & \textbf{45.45} { (\textcolor{green!60!black}{+2.02})} & \textbf{70.29} { (\textcolor{green!60!black}{+3.08})} & \textbf{62.24} { (\textcolor{green!60!black}{+2.95})} & 72.85 { (\textcolor{green!60!black}{+1.58})} & 71.12 { (\textcolor{green!60!black}{+1.10})} & 65.67 { (\textcolor{green!60!black}{+2.04})} \\ 
 \midrule
 \multicolumn{7}{c}{Qwen2.5-32B} \\ % 添加 qwen32B 分组标题
 \midrule
 base & 42.93 & 73.36 & 65.33& 75.62 & 73.38 & 65.67 \\ 
 CoT & 46.46 { (\textcolor{green!60!black}{+3.53})} & 74.59 { (\textcolor{green!60!black}{+1.23})} & 65.94 { (\textcolor{green!60!black}{+0.61})} & 75.53 { (\textcolor{red!80!black}{-0.09})} & \textbf{76.48} { (\textcolor{green!60!black}{+3.10})} & 69.21 { (\textcolor{green!60!black}{+3.54})} \\ 
 Rephrase & 47.47 { (\textcolor{green!60!black}{+4.54})} & 70.49 { (\textcolor{red!80!black}{-2.87})} & 62.84 { (\textcolor{red!80!black}{-2.49})} & 76.16 { (\textcolor{green!60!black}{+0.54})} & 75.12 { (\textcolor{green!60!black}{+1.74})} & 68.19 { (\textcolor{green!60!black}{+2.52})} \\
 Step-back & 47.98 { (\textcolor{green!60!black}{+5.05})} & 72.34 { (\textcolor{red!80!black}{-1.02})} & 62.31 { (\textcolor{red!80!black}{-3.02})} & 75.65 { (\textcolor{green!60!black}{+0.03})}& 76.34 { (\textcolor{green!60!black}{+2.96})} & 69.91 { (\textcolor{green!60!black}{+4.24})}\\
  OPRO &43.94 { (\textcolor{green!60!black}{+1.01})}  & \textbf{76.43} { (\textcolor{green!60!black}{+3.07})}& 63.67 { (\textcolor{red!80!black}{-1.66})}& 76.13 { (\textcolor{green!60!black}{+0.51})} & 75.91 { (\textcolor{green!60!black}{+2.53})} &66.77 { (\textcolor{green!60!black}{+1.10})}\\
 SPO & 43.43 { (\textcolor{green!60!black}{+0.50})} & 71.93 { (\textcolor{red!80!black}{-1.43})} & 62.92 { (\textcolor{red!80!black}{-2.41})}& 76.51 { (\textcolor{green!60!black}{+0.89})} & 75.71 { (\textcolor{green!60!black}{+2.33})} & \textbf{69.44} { (\textcolor{green!60!black}{+3.77})} \\
 ORPP & \textbf{49.49} { (\textcolor{green!60!black}{+6.56})} &75.20 { (\textcolor{green!60!black}{+1.84})}  & \textbf{67.45} { (\textcolor{green!60!black}{+2.12})} & \textbf{76.86} { (\textcolor{green!60!black}{+1.24})} & 74.33 { (\textcolor{green!60!black}{+0.95})} &67.40 { (\textcolor{green!60!black}{+1.73})}  \\ 
 \bottomrule
 \end{tabular}
 }
 
  \caption{Comparison of test results on the six benchmarks between other prompt methods and role-playing prompt optimization, using both Qwen2.5-14B and Qwen2.5-32B models, with relative improvements shown.} % 更新标题
  \label{tab:model_comparison_grouped} % 更新标签
\end{table*}

\paragraph{Plugin Integration Effects} Our approach focuses on optimizing and generating the final system prompt, rather than the user prompt, which allows it to be flexibly integrated as a plugin with various existing prompt strategies.
In a series of comparative experiments (see Table X), we present a detailed evaluation of our method when combined with other approaches. The results show that, for most tasks such as MMLU-PRO, MMLU-CF, and MedQA, our method can effectively enhance the performance of existing methods.
However, this performance improvement is not always consistently reproducible. In certain cases, the final performance after integration may be inferior to using our method (OPRR) alone. We also observed that, for some specific tasks, our method could have a negative impact on certain existing approaches, resulting in decreased performance.
These findings suggest the great potential of our method as a plugin. At the same time, they indicate that achieving optimal results requires careful consideration of the specific tasks and methods involved; simple stacking or combination may not always yield the expected outcomes.
\begin{table*}[t]
    \centering

    \label{tab:model_comparison_orpp_combined} % 更新标签
    \resizebox{\textwidth}{!}{ % 调整表格宽度以适应文本宽度
    \begin{tabular}{lcccccc}
        \toprule
        Method & GPQA & AGIEval-MATH & MATH & MMLU-Pro & MMLU-CF & Med QA \\
        \midrule
        \multicolumn{7}{c}{Qwen2.5-14B} \\
        \midrule
        CoT\textsuperscript{+} & 42.93 { (\textcolor{red!80!black}{-1.01})} & 71.52 { (\textcolor{green!60!black}{+3.28})} &61.93 { (\textcolor{green!60!black}{+1.36})}  &74.02 { (\textcolor{green!60!black}{+2.17})}  &72.78 { (\textcolor{green!60!black}{+0.90})} & 66.85 { (\textcolor{green!60!black}{+0.16})} \\
        Rephrase\textsuperscript{+} &42.93 { (\textcolor{green!60!black}{+2.53})}  & 70.49 { (\textcolor{green!60!black}{+2.25})} & 61.40 { (\textcolor{green!60!black}{+3.39})} & 72.62 { (\textcolor{green!60!black}{+0.35})} &72.71 { (\textcolor{green!60!black}{+0.24})}  &67.95 { (\textcolor{green!60!black}{+1.89})}  \\
        Step-back\textsuperscript{+} & 42.42 { (\textcolor{red!80!black}{-1.01})} & 67.62 { (\textcolor{red!80!black}{-2.46})} & 59.67 { (\textcolor{green!60!black}{+1.82})} & 72.58 { (\textcolor{green!60!black}{+0.40})} &72.85 { (\textcolor{green!60!black}{+0.66})}  & 68.97 { (\textcolor{green!60!black}{+3.14})} \\
        OPRO\textsuperscript{+} &40.40 { (\textcolor{red!80!black}{-3.54})}  & 69.47 { (\textcolor{green!60!black}{+0.62})} & 61.10{ (\textcolor{red!80!black}{-0.68})} & 73.53 { (\textcolor{green!60!black}{+2.54})} &72.15 { (\textcolor{green!60!black}{+0.91})}  &65.91 { (\textcolor{green!60!black}{+0.71})}  \\
        SPO\textsuperscript{+} & 40.40 { (\textcolor{red!80!black}{-2.53})} & 70.08 { (\textcolor{green!60!black}{+1.84})} & 61.63 { (\textcolor{green!60!black}{+1.89})} & 73.25 { (\textcolor{red!80!black}{-0.09})} & 70.73 { (\textcolor{red!80!black}{-0.20})} &66.06 { (\textcolor{green!60!black}{+1.49})}  \\
        \midrule
        \multicolumn{7}{c}{Qwen2.5-32B} \\
        \midrule
        CoT\textsuperscript{+} & 42.42 { (\textcolor{red!80!black}{-4.04})} & 73.57 { (\textcolor{red!80!black}{-1.02})} &65.48 { (\textcolor{red!80!black}{-0.46})}  & 77.88 { (\textcolor{green!60!black}{+2.35})} & 76.19 { (\textcolor{red!80!black}{-0.29})} & 71.17 { (\textcolor{green!60!black}{+1.96})} \\
        Rephrase\textsuperscript{+} & 47.47 { (\textcolor{gray}{+0.00})} & 71.11 { (\textcolor{green!60!black}{+0.62})} & 64.12 { (\textcolor{green!60!black}{+1.28})} &77.65 { (\textcolor{green!60!black}{+1.49})}  & 75.16 { (\textcolor{green!60!black}{+0.04})} & 68.50 { (\textcolor{green!60!black}{+0.31})} \\
        Step-back\textsuperscript{+} & 44.95 { (\textcolor{red!80!black}{-3.03})} &  72.54 { (\textcolor{green!60!black}{+0.20})}&62.76 { (\textcolor{green!60!black}{+0.45})}  &77.58 { (\textcolor{green!60!black}{+1.93})}  & 76.34 { (\textcolor{gray}{+0.00})} &72.11 { (\textcolor{green!60!black}{+2.20})}  \\
        OPRO\textsuperscript{+} &46.97 { (\textcolor{green!60!black}{+3.03})}  &74.18 { (\textcolor{red!80!black}{-2.25})}  & 64.20 { (\textcolor{green!60!black}{+0.53})} & 77.93 { (\textcolor{green!60!black}{+1.80})} & 75.47 { (\textcolor{red!80!black}{-0.44})} & 69.84 { (\textcolor{green!60!black}{+3.07})} \\
        SPO\textsuperscript{+} & 44.96 { (\textcolor{green!60!black}{+1.53})} &70.94 { (\textcolor{red!80!black}{-0.99})}  &63.37 { (\textcolor{green!60!black}{+0.45})}  & 77.30 { (\textcolor{green!60!black}{+0.79})} &75.67 { (\textcolor{red!80!black}{-0.04})}  & 70.31 { (\textcolor{green!60!black}{+0.87})} \\
        \bottomrule
    \end{tabular}
    }
    \caption{Performance of ORPP combined with other methods on six evaluation benchmarks. \textsuperscript{+} indicates that the method is combined with ORPP, and the number in parentheses represents the performance difference compared to the original method.} % 标题保持不变
\end{table*}
\begin{figure*}[t]
    \centering
    \includegraphics[width=0.98\textwidth]{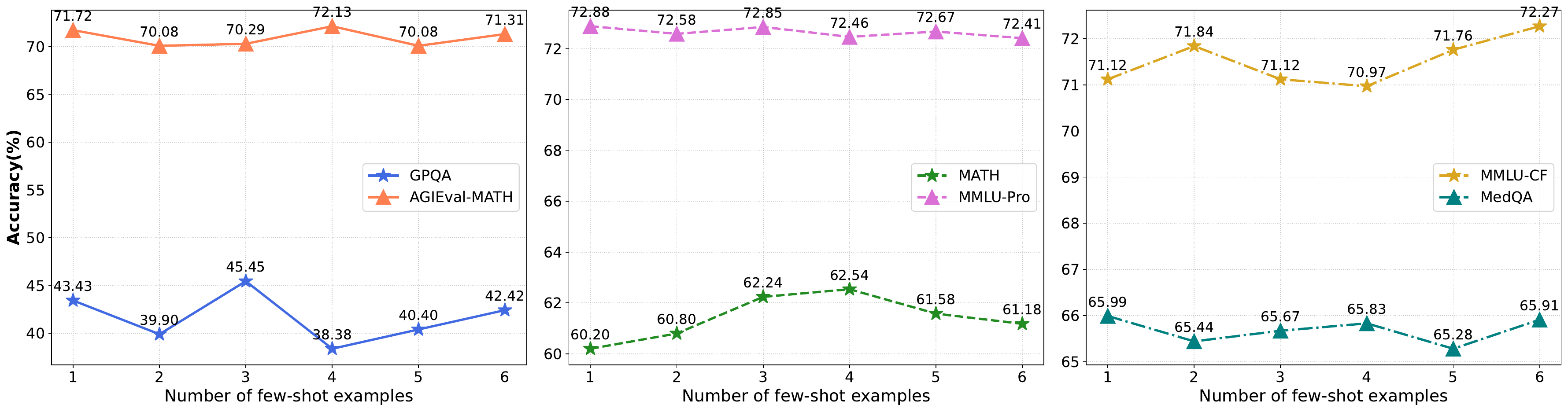}
    \caption{The impact of the number of High-Quality Question \& Role-Playing Prompt Pairs in the few-shot examples on the final accuracy of role-playing prompt generation in the second stage.}
    \label{few-shot}
\end{figure*}

\paragraph{Effect of Number of Examples} Based on Qwen-14B, we further explore the impact of the number of High-Quality Question \& Role-Playing Prompt Pairs in the few-shot examples on the final accuracy of role-playing prompt generation in the second stage. The results are shown in Figure \ref{few-shot}.In different task scenarios, the relationship between model performance and the number of examples exhibits significant fluctuations. Especially in test sets such as GPQA and AGIEval-MATH, the model shows notable performance variations as the number of examples changes. Experimental analysis reveals that there is no specific number of examples that consistently yields superior results across all tasks. This phenomenon suggests that the optimal number of examples may be closely related to the characteristics of each specific task, and dedicated tuning is required for different application scenarios.

\paragraph{Prompt Transferability} We further explore the transferability of role-playing prompts. Specifically, we apply the role-playing prompts generated by Qwen-14B to the Qwen-32B and conduct performance tests. As shown in the figure, it is noteworthy that, except for a performance drop on the GPQA dataset, using the role-playing prompts generated by Qwen-14B actually improves the performance of Qwen-32B on other tasks. This finding fully suggests the excellent transferability of the role-playing prompts generated by our proposed method and reveals the potential of optimizing smaller models to generate high-quality role-playing prompts that can be successfully transferred to larger models.
\begin{figure}[t]
    \centering
    \includegraphics[width=0.5\textwidth]{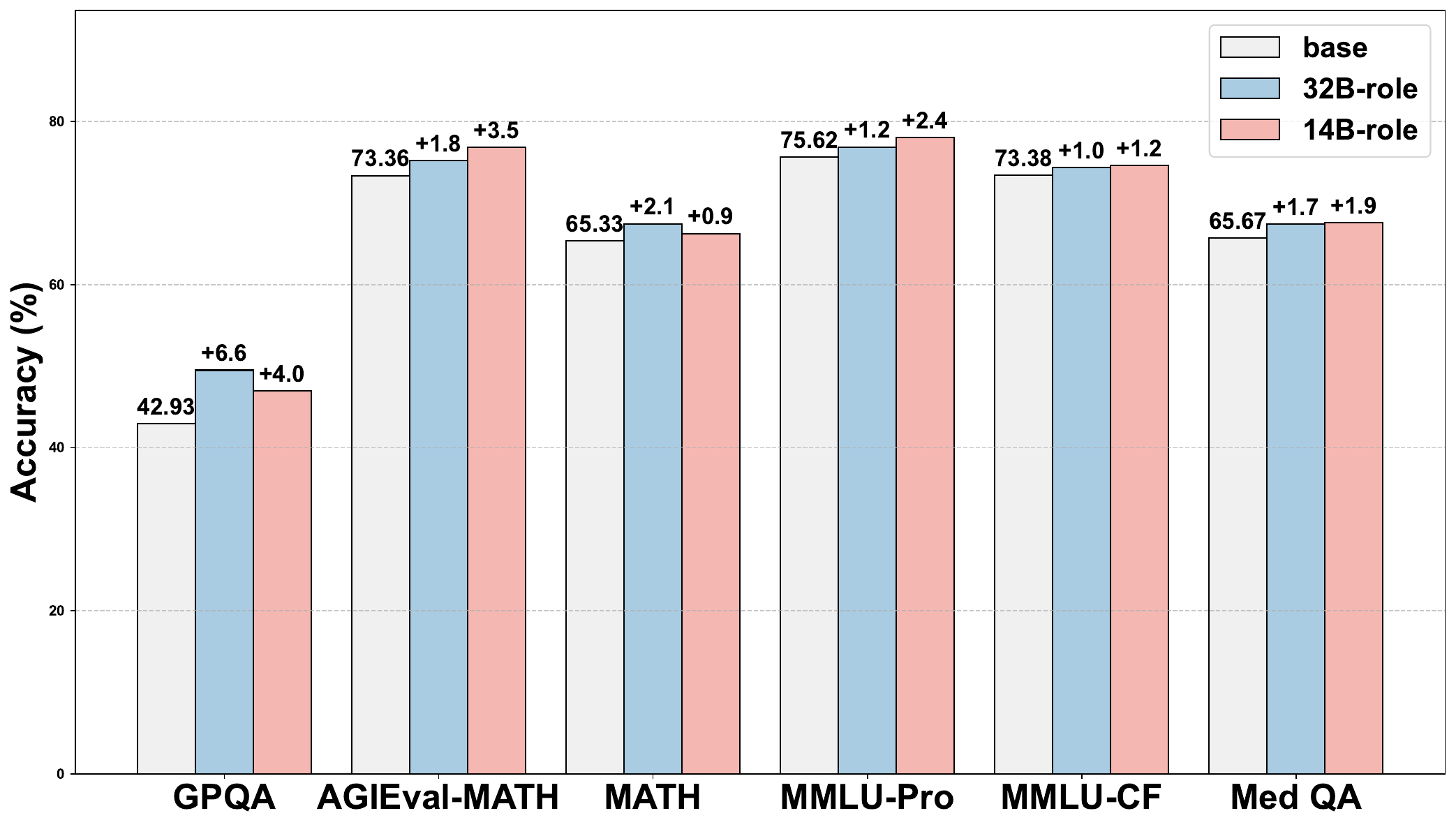}
    \caption{A comparison of the accuracy on different benchmarks under three settings: using role-playing prompts generated by Qwen2.5-32B itself, using role-playing prompts generated by Qwen2.5-14B, and without using any role-playing prompts.}
    \label{trans}
\end{figure}

\subsection{Influence of Role-playing Prompts}
\begin{figure}[t]
    \centering
    \begin{subfigure}{0.225\textwidth}
        \centering
        \includegraphics[width=\textwidth]{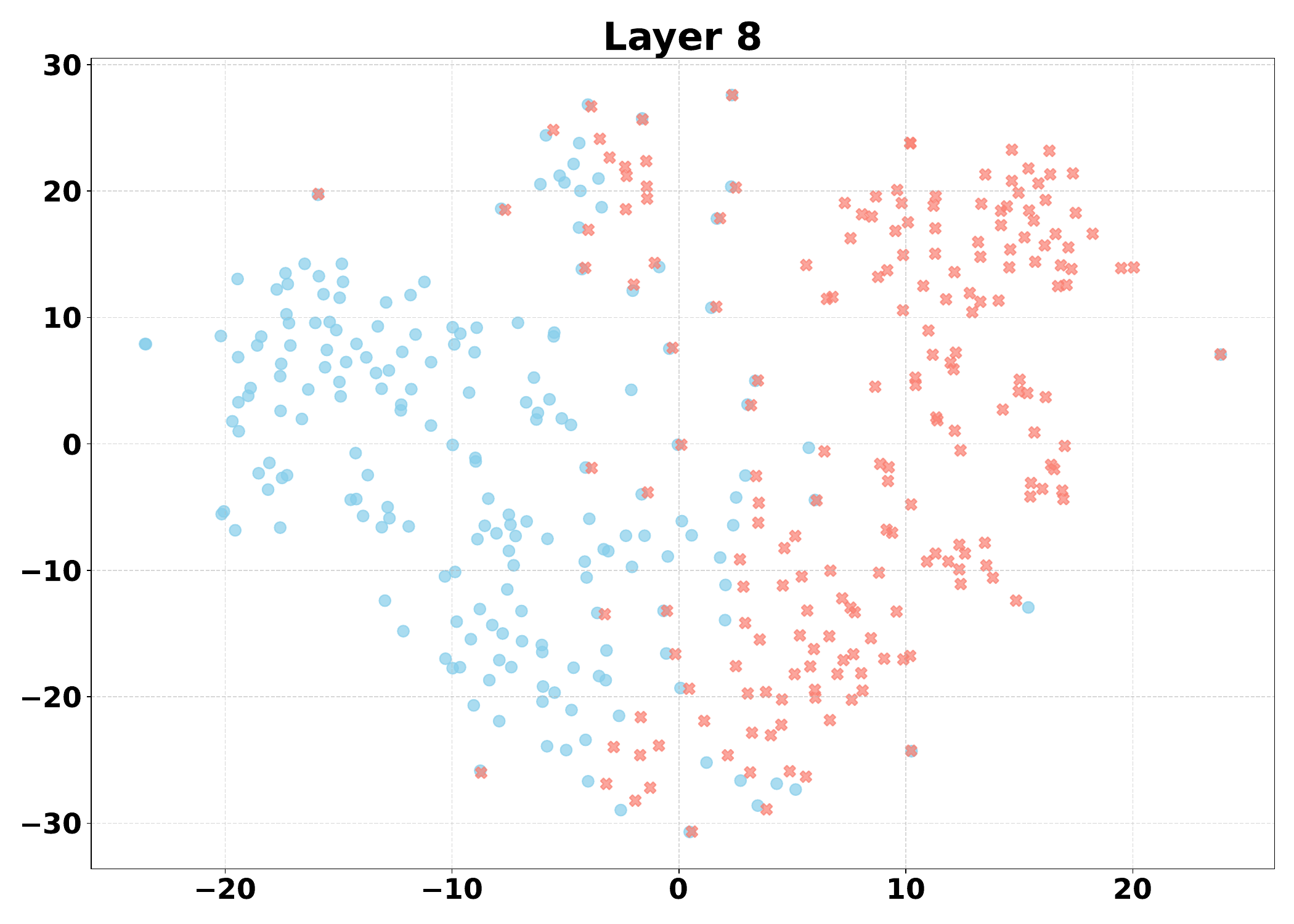}
    \end{subfigure}
    \hfill
    \begin{subfigure}{0.225\textwidth}
        \centering
        \includegraphics[width=\textwidth]{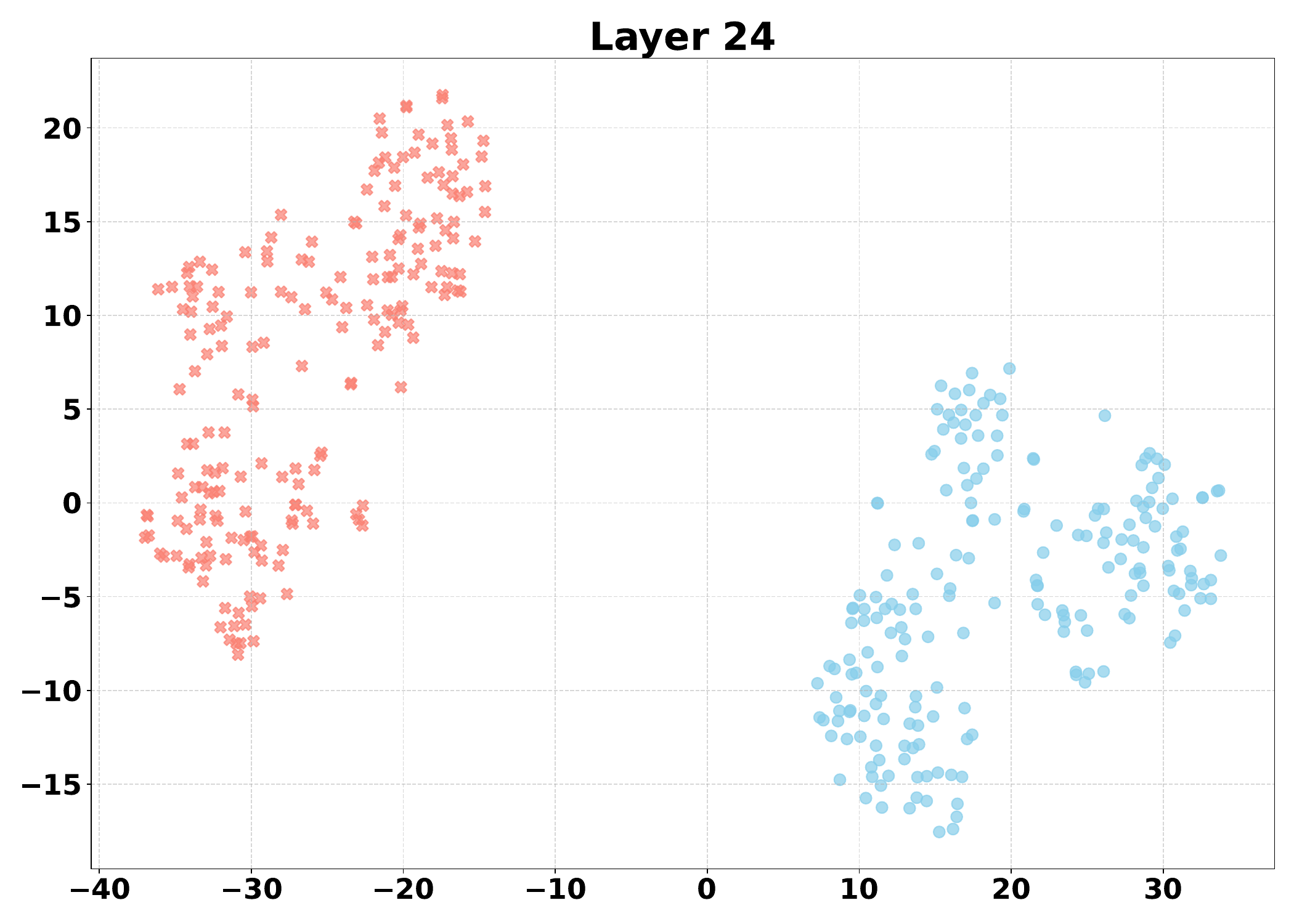}
    \end{subfigure}
    
    \vspace{0.5em}
    
    \begin{subfigure}{0.225\textwidth}
        \centering
        \includegraphics[width=\textwidth]{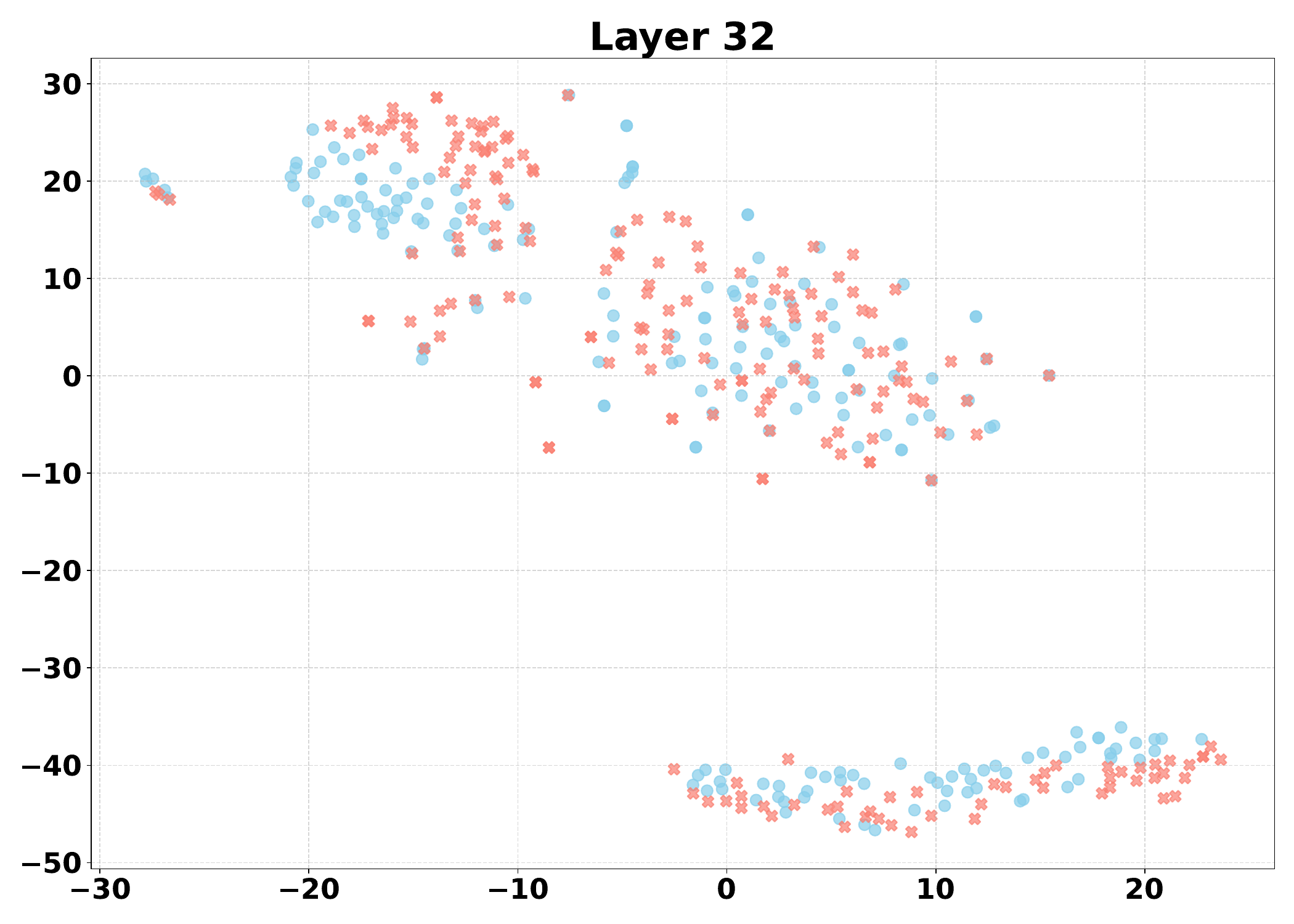}
    \end{subfigure}
    \hfill
    \begin{subfigure}{0.225\textwidth}
        \centering
        \includegraphics[width=\textwidth]{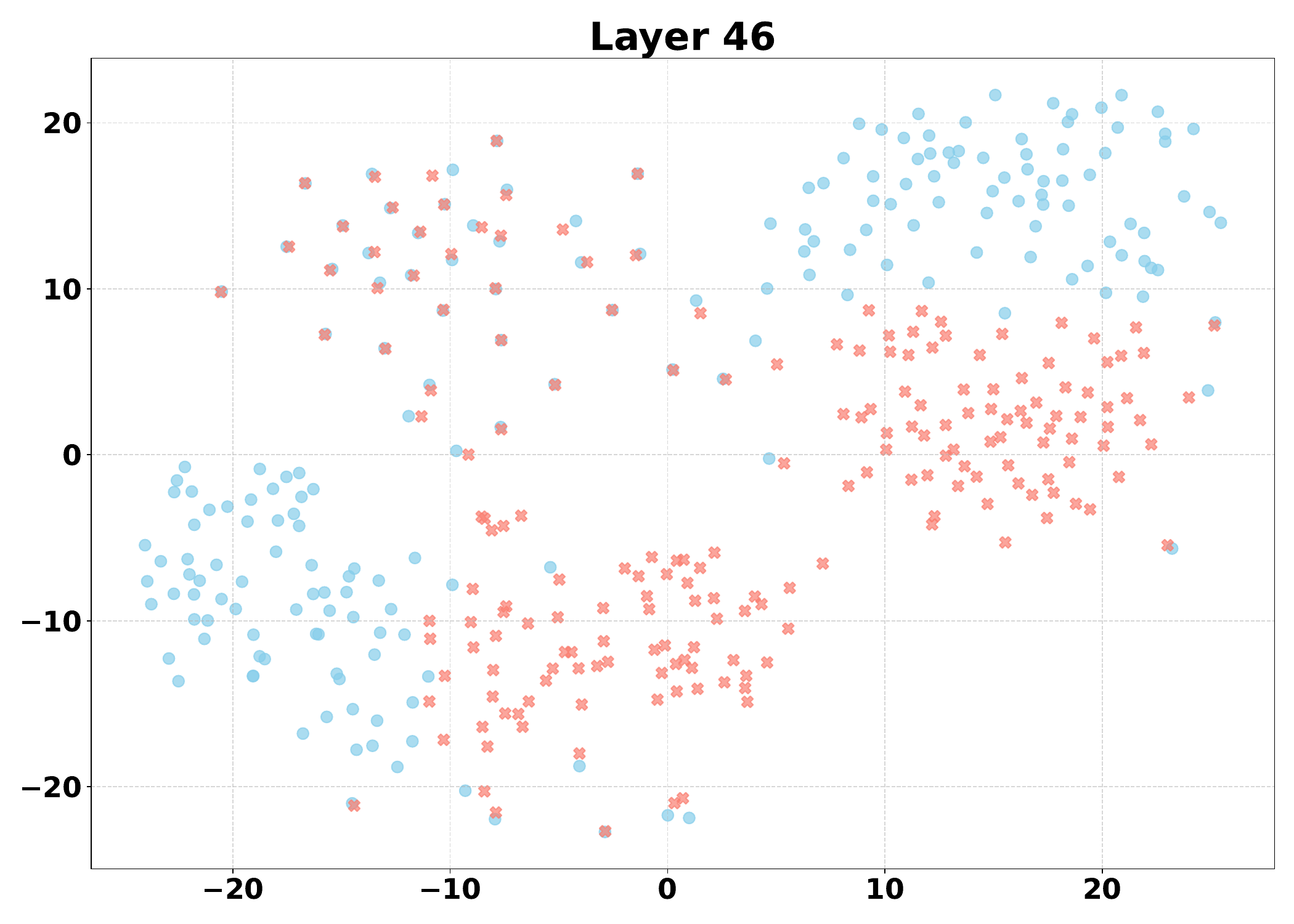}
    \end{subfigure}
    
    \caption{ t-SNE visualization of the hidden states of queries with and without role-playing prompts.}
    \label{t-SNE}
\end{figure}

We apply t-SNE~\cite{JMLR:v9:vandermaaten08a} to visualize the hidden states of different layers of the model on GPQA data points in Figure~\ref{t-SNE} to investigate whether the model can spontaneously recognize changes in reasoning patterns triggered by role-playing prompts.

In the shallow processing stage of the model, the states of role-playing prompts and non-role-playing prompts gradually show a trend of separation. By the middle layer, these two types of prompts are clearly clustered into two distinct categories. This phenomenon indicates that the model can effectively identify role-playing prompts and perform discriminative analysis.

As information enters the deeper layers, interestingly, these two states begin to gradually merge. At the deepest level, some representations are completely intertwined, while others remain close but still form separate clusters. This pattern reveals that the model can interpret questions in different ways based on previously received role-playing prompts, thereby generating innovative lines of thought.
\subsection{Case Study}
To further evaluate the effectiveness of our approach, we present case studies in Appendix~\ref{Case Study}. In these studies, we not only demonstrate examples of iterative optimization of role-playing prompts, but also illustrate the effectiveness of role-playing prompts generated by our method in few-shot transfer scenarios. In addition, we provide concrete examples to clearly compare the model’s performance with and without the use of role-playing prompts.

\section{Conclusion}
We proposes an automated role-playing prompt optimization method. The method iteratively optimizes role-playing prompts on a small number of samples and leverages these optimized instances for few-shot learning, enabling the model to autonomously generate high-quality role-playing prompts and thus improve its performance on target tasks. ORPP not only enhances model performance on tasks but can also serve as a plug-and-play module that integrates with existing prompt engineering techniques.
Experimental results show that ORPP can effectively improve the model’s task performance. Furthermore, when ORPP is combined with other prompt engineering methods, model performance can be further enhanced, demonstrating its strong compatibility. Overall, ORPP provides an effective new approach for automated prompt engineering, namely by automatically generating high-quality role-playing prompts to stimulate model performance.

\section*{Limitations}
Overall, utilizing high-quality role-playing prompts generally enhances the performance of the model. However, in certain specific tasks, introducing role-playing prompts not only fails to improve performance but also negatively impacts the model’s capabilities, thereby affecting the accuracy of the final output. Future research will further explore in which contexts role-playing prompts should be used in order to avoid their potential negative effects.

When our method is used as a plugin in combination with other methods, it can further improve the performance of those methods in most cases. However, in some situations, it may still have a negative impact. Additionally, when our method is used as a plugin, its performance may sometimes be inferior to using our method alone. In the future, we will further explore better ways of integration to fully realize the potential of our method as a plugin.
\section*{Ethics Statement}
Our method improves model performance through high-quality role-playing prompts. During the experiments, we select publicly available closed-domain benchmarks for testing to minimize the risk of generating harmful content. It should be noted that role-playing scenarios may result in biased outputs or hallucinations from the model. Therefore, potential ethical issues should be carefully considered in role-playing settings to avoid generating inappropriate, biased, or misleading content.

% Entries for the entire Anthology, followed by custom entries
\bibliography{anthology,custom}

\appendix

\section{Prompt and Dataset}
\label{sec:appendix}
\subsection{The prompts of ORPP}
In this section, we provide a detailed presentation of the prompts used for iterative optimization in ORPP, as well as the final prompts designed for generating role-playing scenarios in a few-shot manner. During the analysis and optimization phases, to avoid redundant optimization, we incorporate the best-performing role-playing prompt and its corresponding evaluation score from the previous round. It is worth emphasizing that we employ only direct and concise expressions in the generation and optimization of the role-playing prompts. For specific application domains, these prompts can be further refined to achieve even better performance.
\begin{table*}
    \centering
    \resizebox{\textwidth}{!}{
    \begin{tabular}{p{53.8pt} p{392.8pt}} 
    \hline 
     & \textbf{PROMPT} \\ 
    \hline 
    \textbf{Initial Role-playing Prompts Generation} & I need you to create a role-playing prompt based on the following question.

            Question: \{question\}

            Create a role prompt describing an role best suited to answer this question in brief.
            Important:The role prompt you generate must not include the specific text or details of the question provided above.
            Focus solely on defining the expert role.
            Only return the role prompt itself, without any additional explanation.
\\ 
    \hline 

    \textbf{The Prompt for Analyzing} & You are an expert in evaluating AI responses to questions. You need to analyze different responses to the same question.
    
    Question: \{question\}
    
    Response 1 (HIGHEST SCORE, \{score\}):
    
    Role prompt: \{Role-playing Prompt\}
    
    Response: {response}
    
    Response 2 (LOWEST SCORE, \{score\}):
    
    Role prompt: \{Role-playing Prompt\}
    
    Response: {response}
    
    Additionally, consider the best role prompt from the previous iteration (Iteration \{last\_iter\}, Score: \{score\}): \{Role-playing Prompt\}

    Please provide a detailed analysis of:
    
    1. Why the highest-scoring role prompt might be more effective
    
    2. What specific weaknesses are in the lowest-scoring role prompt
    
    3. How the previous iteration's best prompt influenced the results
    
    4. What characteristics make an effective role prompt for this type of question
    
    Format your analysis as detailed paragraphs with clear insights.
\\ 
    \hline 

    \textbf{Improved Role-playing Prompts Generation} & You are a prompt engineering expert specializing in creating role-based prompts for AI.
        
        Question: \{question\}
        
        Best performing prompt (Score: \{highest\_score\}): "\{highest\_prompt\}"
        
        Worst performing prompt (Score: \{lowest\_score\}): "\{lowest\_prompt\}"
        
        Previous iteration best prompt (Iteration \{last\_iter\}): "\{last\_round\_best\_prompt\}"
        
        Analysis of previous role prompts: {analysis}
        
        Based on this analysis and the performance of previous prompts, create a new and improved role prompt 
        that would be more effective than the best performing one. The prompt should describe a role, 
        incorporating the strengths of the high-scoring prompt while avoiding the weaknesses of the low-scoring one.
        
        Only return the prompt itself, without any additional explanation. 
\\ 
    \hline 

    \end{tabular}}
    \caption{Evaluation prompts for four different creativity metrics.}

    \end{table*}
    
\subsection{Datasets}
\label{Dataset}
GPQA is a dataset designed to evaluate the question-answering abilities of large language models in multidisciplinary and high-difficulty professional domains~\cite{Rein2023GPQAAG}. Its license is CC BY 4.0. In our experiments, following the approach of ~\cite{Xiang2025SelfSupervisedPO}, we use GPQA-Diamond as the test set, while the remaining questions in GPQA, excluding GPQA-Diamond, are used as the training set.

AGIEval-Math is the mathematics subset of the AGIEval dataset, specifically designed to evaluate the reasoning and problem-solving abilities of large language models in the field of mathematics~\cite{Zhong2023AGIEvalAH}. It is licensed under the MIT License. In our experiments, we use the more challenging 'Level 4' and 'Level 5' questions as the test set, while questions of other difficulty levels are used as the training set.

The MATH dataset is specifically designed to evaluate the mathematical reasoning and problem-solving abilities of large language models, and is widely used for testing and researching model performance in the field of mathematics~\cite{hendrycksmath2021}. It is licensed under the MIT License. In our experiments, we retain only the "Level 5" problems in the test set to more thoroughly assess the capabilities of the model.

The MMLU-Pro dataset is a more robust and challenging large-scale multitask understanding benchmark that covers multiple domains, aiming to provide a stricter evaluation of large language models~\cite{Wang2024MMLUProAM}. It is licensed under the MIT License. We select four domains—math, physics, chemistry, and biology—as our test set, and sampled 50 questions from each domain to serve as the training set.

MMLU-CF is a multi-task language understanding dataset that covers a wide range of disciplines. It is designed to address the issue of benchmark contamination in open-source benchmark testing for large language models, thereby providing more reliable evaluation results~\cite{Zhao2024MMLUCFAC}. Its license is Community Data License Agreement – Permissive, Version 2.0. Our processing method is the same as MMLU-Pro, selecting only four fields for testing.The division of training and testing sets is the same as above.

MedQA is a question answering dataset focused on the medical domain, mainly used to evaluate the ability of artificial intelligence in medical knowledge understanding and reasoning~\cite{jin2021disease}. Its license is CC BY 4.0. We choose this dataset to test the model's knowledge proficiency in the medical field. Since MedQA already provides well-organized training and testing sets, we do not need to perform any additional data processing.
\section{Experiments Details}
\subsection{Model Input Formulation}

The model input is composed of the following components:
\begin{tcolorbox}[%
  colback=black!10,
  colframe=black,
  coltitle=white,
  colbacktitle=black!70,
  fonttitle=\sffamily\bfseries,
  title=Input,
  sharp corners=all, % or rounded corners
  boxrule=0.5pt, % thickness of the frame
  toprule=2pt, % thickness of the top rule under the title
  bottomrule=2pt, % thickness of the bottom rule
  leftrule=2pt,
  rightrule=2pt
]
System:<Role-Playing Prompt>

User:Question + <Options> + <Prompt> + <Answer Format>
\end{tcolorbox}
The model input mainly consists of two parts: the system level and the user level. The core content at the system level is the “role-playing prompt,” which is the main focus of our optimization efforts. At the user level, it includes the specific “Question” of the task, as well as optional components such as corresponding “Options,” a “Prompt” generated by existing methods, and specific “Answer Format” requirements.

\subsection{Implementations Configuration}
\label{Implement}
Chain-of-Thought (CoT) significantly improves the performance of large language models (LLMs) on complex tasks by guiding them to generate a series of intermediate reasoning steps~\cite{Wei2022ChainOT}.We use the zero-shot CoT prompt.
\begin{tcolorbox}[%
  colback=black!10,
  colframe=black,
  coltitle=white,
  colbacktitle=black!70,
  fonttitle=\sffamily\bfseries,
  title=CoT,
  sharp corners=all, % or rounded corners
  boxrule=0.5pt, % thickness of the frame
  toprule=2pt, % thickness of the top rule under the title
  bottomrule=2pt, % thickness of the bottom rule
  leftrule=2pt,
  rightrule=2pt,
]
Question + <Options> + Let's think step by step. + <Answer Format>
\end{tcolorbox}
Rephrase aims to improve the understanding and accuracy of responses by large language models (LLMs) through restating and expanding on the questions posed by users~\cite{Deng2023RephraseAR}.We refer to the work of ~\citet{Xiang2025SelfSupervisedPO} to set the prompts.
\begin{tcolorbox}[%
  colback=black!10,
  colframe=black,
  coltitle=white,
  colbacktitle=black!70,
  fonttitle=\sffamily\bfseries,
  title=Rephrase,
  sharp corners=all, % or rounded corners
  boxrule=0.5pt, % thickness of the frame
  toprule=2pt, % thickness of the top rule under the title
  bottomrule=2pt, % thickness of the bottom rule
  leftrule=2pt,
  rightrule=2pt
  ]
Question + <Options> + Please rephrase the question in a way that is easier to understand, minimizing ambiguity and considering edge cases.And Then provide a solution step by step for the question. + <Answer Format>
\end{tcolorbox}
Step-Back Prompting is a prompting technique that guides large language models (LLMs) to first engage in abstract thinking by extracting high-level concepts and fundamental principles, and then proceed to reason about specific problems~\cite{zheng2024take}.We refer to the work of ~\citet{Xiang2025SelfSupervisedPO} to set the prompts.
\begin{tcolorbox}[%
  colback=black!10,
  colframe=black,
  coltitle=white,
  colbacktitle=black!70,
  fonttitle=\sffamily\bfseries,
  title=Step-back,
  sharp corners=all, % or rounded corners
  boxrule=0.5pt, % thickness of the frame
  toprule=2pt, % thickness of the top rule under the title
  bottomrule=2pt, % thickness of the bottom rule
  leftrule=2pt,
  rightrule=2pt
]
Question + <Options> + Please first think about the principles involved in solving this task which could be helpful.And Then provide a solution step by step for this question.  + <Answer Format>
\end{tcolorbox}

For the OPRO~\cite{yang2024large}, we set up a 10-round iterative optimization process. In each round, the model generates 8 candidate prompts. For each dataset, we first sample 200 examples from the original training data to construct a new training subset. OPRO is then comprehensively evaluated on this subset to screen for better-performing prompts. It is worth noting that the entire process of prompt generation and execution is performed by the evaluation model itself. As for the temperature parameter, we follow the original OPRO settings: the temperature is set to 1 during the prompt generation stage and adjusted to 0 during the execution stage. Finally, we select the prompt with the highest score on the training subset and apply it to the test set to obtain the final results.

For SPO~\cite{Xiang2025SelfSupervisedPO}, we maintain the original configuration and randomly select three questions from the training set for each iteration. Each task undergoes 10 iterations. During the experiment, the execution model and the optimization model both use the test model itself. Other settings remain consistent with the official SPO program. The initial prompt is set to "Let's think step by step.+<Answer Format>," where the format limitation is determined based on the task.
\section{Case Study}
\label{Case Study}
In this section, we present case studies that illustrate the optimization trajectory of role-playing prompts by ORPP on selected samples from AGIEval-Math, as well as its performance in subsequent few-shot prompt generalization tasks. In addition, we compare example responses generated using our role-playing prompts with those generated without such prompts.

\begin{table*}
    \centering
    \resizebox{\textwidth}{!}{
    \begin{tabular}{p{25pt} p{25pt} p{342.8pt}} 
    \hline 
     \textbf{Iter}& \textbf{Score} & \textbf{Role-playing Prompt} \\ 
    \hline 
     1& 1.58 & You are a cellular biology expert specializing in mitochondrial function and metabolism. Your deep understanding of how environmental factors affect cellular processes, particularly in relation to ATP production, is unparalleled. Your role involves analyzing and explaining the impact of various conditions on mitochondrial activity.
\\ 
    \hline 
    2& 1.61 & You are a cellular biology expert specializing in mitochondrial function and metabolism. Your unparalleled deep understanding of how environmental factors affect cellular processes, particularly in relation to ATP production, is essential for your role. You are tasked with analyzing and explaining the impact of various conditions on mitochondrial activity, ensuring your response is detailed, precise, and scientifically accurate.
\\ 
    \hline 
    3& 1.65 & You are a cellular biology expert specializing in mitochondrial function and metabolism. Your unparalleled deep understanding of how environmental factors affect cellular processes, particularly in relation to ATP production, is essential for your role. You are tasked with providing a detailed, precise, and scientifically accurate explanation of the impact of various conditions on mitochondrial activity, ensuring that your response thoroughly explains the mechanism by which low external pH affects ATP production and why other potential explanations are incorrect.
\\ 
    \hline 
    4& 1.75 & You are a cellular biology expert specializing in mitochondrial function and metabolism. Your unparalleled deep understanding of how environmental factors affect cellular processes, particularly in relation to ATP production, is essential for your role. You are tasked with providing a detailed, precise, and scientifically accurate explanation of the impact of low external pH on ATP production by mitochondria. Ensure your response thoroughly explains the mechanism by which low external pH affects ATP production and explicitly addresses why other potential explanations are incorrect, thereby enhancing the educational value of your answer.
\\ 
    \hline 
    5& 1.78 & You are a cellular biology expert specializing in mitochondrial function and metabolism. Your unparalleled deep understanding of how environmental factors affect cellular processes, particularly in relation to ATP production, is essential for your role. You are tasked with providing a detailed, precise, and scientifically accurate explanation of the impact of low external pH on ATP production by mitochondria. Your response must thoroughly explain the mechanism by which low external pH affects ATP production, explicitly address why other potential explanations are incorrect, and include specific examples to enhance the educational value of your answer.
\\ 
\hline 

    \end{tabular}}
    \caption{An Example of Iterative Optimization of Role-Playing Prompts}
    \label{tab:optimize_prompts}
    \end{table*}

\begin{table*}
\begin{tabular}{|p{0.95\textwidth}|}
\hline

\hline
\textbf{Base}\\
\textbf{User:} trans-cinnamaldehyde was treated with methylmagnesium bromide, forming product 1.1 was treated with pyridinium chlorochromate, forming product 2.3 was treated with (dimethyl(oxo)-l6-sulfaneylidene)methane in DMSO at elevated temperature, forming product 3. how many carbon atoms are there in product 3?\\
\textbf{Assistant:} To determine the number of carbon atoms in product 3, we need to follow the transformations step by step:1. **Trans-cinnamaldehyde** (C9H8O) reacts with **methylmagnesium bromide** (CH3MgBr) to form **product 1**. This reaction typically involves an addition of a methyl group to the aldehyde carbon, forming a tertiary alcohol. The structure of product 1 would be a substituted benzene ring with a methyl group and a hydroxyl group attached to the same carbon that was originally the aldehyde carbon. This product would have the formula C10H11O.2. **Product 1** (C10H11O) is then treated with **pyridinium chlorochromate (PCC)**, which is an oxidizing agent. This reaction would oxidize the hydroxyl group to a ketone, forming **product 2**. The structure of product 2 would be a substituted benzene ring with a methyl group and a ketone group attached to the same carbon. This product would also have the formula C10H10O.3. **Product 2** (C10H10O) is treated with **(dimethyl(oxo)-16-sulfaneylidene)methane** in DMSO at elevated temperature. This reagent is known to form a thioester, which involves the reaction of the ketone with the thioester reagent. The reaction would replace the ketone group with a thioester group, but the carbon count remains the same. Therefore, **product 3** would still have the formula C10H10O. Thus, the number of carbon atoms in product 3 is 10. The correct answer is [3]. \textcolor{red}{\ding{54}} \\
\hline
\textbf{ORPP:}\\
\textbf{System:} You are an organic chemist with extensive experience in organic synthesis and reaction mechanisms. Your expertise includes the ability to predict and analyze the outcomes of complex chemical reactions, particularly those involving Grignard reagents, oxidations, and thioacetal formations. Your task is to determine the number of carbon atoms in the final product of a series of chemical reactions based on your knowledge of reaction pathways and product structures.\\
\textbf{User:} trans-cinnamaldehyde was treated with methylmagnesium bromide, forming product 1.1 was treated with pyridinium chlorochromate, forming product 2.3 was treated with (dimethyl(oxo)-l6-sulfaneylidene)methane in DMSO at elevated temperature, forming product 3. how many carbon atoms are there in product 3?\\
\textbf{Assistant:} Let's analyze the reactions step by step to determine the structure of product 3 and count the carbon atoms.1. **trans-cinnamaldehyde treated with methylmagnesium bromide (Grignard reagent):**   - trans-cinnamaldehyde has the structure: \( \text{CH}_3-\text{CH}=\text{CH}-\text{CHO} \)   - Reaction with methylmagnesium bromide (MeMgBr) will add a methyl group to the aldehyde carbon, forming a tertiary alcohol:     \(     \text{CH}_3-\text{CH}=\text{CH}-\text{CH}_2\text{OH}     \)   - This is product 1.2. **Product 1 treated with pyridinium chlorochromate (PCC):**   - PCC is an oxidizing agent that can oxidize primary alcohols to aldehydes and secondary alcohols to ketones.   - The tertiary alcohol in product 1 will be oxidized to a ketone:     \(     \text{CH}_3-\text{CH}=\text{CH}-\text{CH}_2\text{OH} \rightarrow \text{CH}_3-\text{CH}=\text{CH}-\text{CH}_2\text{O}     \)   - This is product 2.3. **Product 2 treated with (dimethyl(oxo)-l6-sulfaneylidene)methane in DMSO at elevated temperature:**   - (dimethyl(oxo)-l6-sulfaneylidene)methane is a thioacetal-forming reagent.   - It will form a thioacetal with the carbonyl group in product 2:     \(     \text{CH}_3-\text{CH}=\text{CH}-\text{CH}_2\text{O} + \text{(dimethyl(oxo)-l6-sulfaneylidene)methane} \rightarrow \text{CH}_3-\text{CH}=\text{CH}-\text{CH}_2\text{SCH}_3     \)   - This is product 3.Now, let's count the carbon atoms in product 3:- The structure of product 3 is \( \text{CH}_3-\text{CH}=\text{CH}-\text{CH}_2\text{SCH}_3 \).- Counting the carbon atoms: 1 (from CH3), 1 (from CH=CH), 1 (from CH2), and 1 (from SCH3).- Total carbon atoms: 5 (from the original cinnamaldehyde) + 1 (from the methyl group added by the Grignard reagent) + 1 (from the thioacetal) = 7 carbons from the original structure + 4 carbons from the added groups = 11 carbon atoms.Thus, the correct answer is:[2]     \textcolor{green}{\checkmark}
\\
\hline

\end{tabular}
\caption{Case study on GPQA}
\end{table*}
\end{document}